\documentclass{preprint}

\usepackage[colorlinks=true]{hyperref}
\usepackage[numbers,comma,sort&compress]{natbib} 
\usepackage{listings}
\usepackage{graphicx}
\usepackage{booktabs}       
\usepackage{xcolor}
\usepackage{mdframed}
\usepackage{xspace}
\usepackage{multirow}
\usepackage{subcaption}
\mdfdefinestyle{prompt}{%
    linecolor=black,
    outerlinewidth=2pt,
    innertopmargin=4pt,
    innerbottommargin=4pt,
    innerrightmargin=4pt,
    innerleftmargin=4pt,
    leftmargin = 0pt,
    rightmargin = 0pt,
    backgroundcolor=gray!5
}

\newcommand{\factchecker}{\textsc{FactChecker}\xspace}
\title{Enhancing Health Fact-Checking with LLM-Generated Synthetic Data}
\author[1]{Jingze Zhang}
\author[1]{Jiahe Qian}
\author[2]{Yiliang Zhou}
\author[1,*]{Yifan Peng}
\affil[1]{Population Health Sciences, Weill Cornell Medicine, New York, 10022, US}
\affil[2]{Donald Bren School of Information and Computer Sciences, University of California, Irvine, Irvine, 92697, US}
\affil[*]{Corresponding author(s). Email(s): \url{yip4002@med.cornell.edu}}

\begin{document}

\maketitle

\begin{abstract}
  Fact-checking for health-related content is challenging due to the limited availability of annotated training data. In this study, we propose a synthetic data generation pipeline that leverages large language models (LLMs) to augment training data for health-related fact checking. In this pipeline, we summarize source documents, decompose the summaries into atomic facts, and use an LLM to construct sentence–fact entailment tables. From the entailment relations in the table, we further generate synthetic text–claim pairs with binary veracity labels. These synthetic data are then combined with the original data to fine-tune a BERT-based fact-checking model. Evaluation on two public datasets, PubHealth and SciFact, shows that our pipeline improved F1 scores by up to 0.019 and 0.049, respectively, compared to models trained only on the original data. These results highlight the effectiveness of LLM-driven synthetic data augmentation in enhancing the performance of health-related fact-checkers.
\end{abstract}



\maketitle

\section{Introduction}

Fact-checking is the process of evaluating the truthfulness of claims by systematically comparing them against credible and authoritative sources of evidence \cite{kotonya-toni-2020-explainable-automated,vlachos-riedel-2014-fact}. This process is crucial to combating misinformation, especially in public health, where false or misleading medical information can directly harm public well being \cite{waszak2018spread}. Ensuring that health‐related claims are accurate is not only fundamental to upholding scientific integrity, but also essential for preserving public trust.

One key challenge in developing reliable fact-checkers in medicine is the lack of labeled training data. 
Fact-checking in the health domain requires specialized medical expertise, making the annotation process costly and time-consuming \cite{kazari2025scaling}. As a result, models trained on general-purpose fact-checking corpora often struggle to generalize to medical claims, highlighting the critical need for more domain-specific data. However, existing datasets have notable limitations when applied to health-related contexts. For instance, the widely used FEVER dataset contains over 185,000 synthetic claims derived from Wikipedia, but it does not require domain knowledge and therefore lacks applicability to medical fact-checking tasks \cite{thorne2018fever}. While PubHealth offers approximately 11,000 journalist-verified claims focused on public health, its topic coverage is narrow and biased toward false statements \cite{kotonya-toni-2020-explainable-automated}. Similarly, SciFact is tailored to verifying scientific claims against paper abstracts but comprises only about 1,400 annotated instances—an insufficient quantity for training large language models \cite{wadden-etal-2020-fact}.

% Few existing datasets are adequate for training and evaluating fact‐checkers on complex health‐related claims.
%Our work presents a method for synthesizing fact-checking instances from arbitrary documents, enabling the training of fact-checking models.
% Recent work has also shown that large language models (LLMs) can verify claims in zero- and few-shot settings, reducing reliance on large-labeled corpora. One study demonstrated that a generative LLM can generate and verify the entailed versus contradictory claim–document pairs using synthetic examples, which improves downstream fact-checking performance \cite{choi2024factgpt}. Another approach proposed a logical reasoning framework for LLMs, empirically validating their capacity to assess claim veracity with minimal fine-tuning \cite{dougrezlewis2024assessing}. Although these studies indicate that LLMs possess inherent reasoning capabilities for fact checking, their large size presents challenges for local deployment in healthcare settings, often involving privacy and security concerns.

In this study, we introduce a synthetic data generation pipeline for fact‐checking. We show that, by leveraging synthetic data, it is possible to fine-tune a smaller model (e.g., BERT) to effectively verify multiple facts within grounding documents. Specifically, the data synthesis pipeline first summarizes each document and extracts its atomic facts while simultaneously splitting the original text into individual sentences. A large language model (LLM) is then used to construct a sentence–fact support table, marking whether each sentence supports each fact. To generate synthetic data, we sample a subset of sentences, select one of the extracted facts as a synthetic claim, and automatically assign its veracity label by consulting the support table. Finally, we merge these synthetic examples with the original data to fine-tune a BERT-based fact-checking model, thus creating a richer training set.  Specifically, our approach led to F1 score improvements of up to 0.019 on PubHealth and 0.049 on SciFact compared to models trained only on the original data.

Our contributions are as follows: (1) We propose an LLM-driven pipeline that constructs claim–text pairs from arbitrary documents using atomic facts, (2) We implement a BERT-based \factchecker using the synthetic data, (3) Evaluation on the PubHealth and SciFact datasets shows that, by leveraging synthetic data, it is possible to fine-tune a smaller model (e.g., BERT) to effectively verify multiple facts within grounding documents.

% \section{Related work}

% \paragraph{Challenges in health-related fact-checking due to dataset gaps}

% \paragraph{Fine-tuning small models with synthetic data}

%teach a small model how to simultaneously verify multiple facts in a sentence against multiple sentences in grounding documents.

\section{Methodology}

\subsection{Overview}

\begin{figure}[t]
\centering
\includegraphics[width=0.99\textwidth]{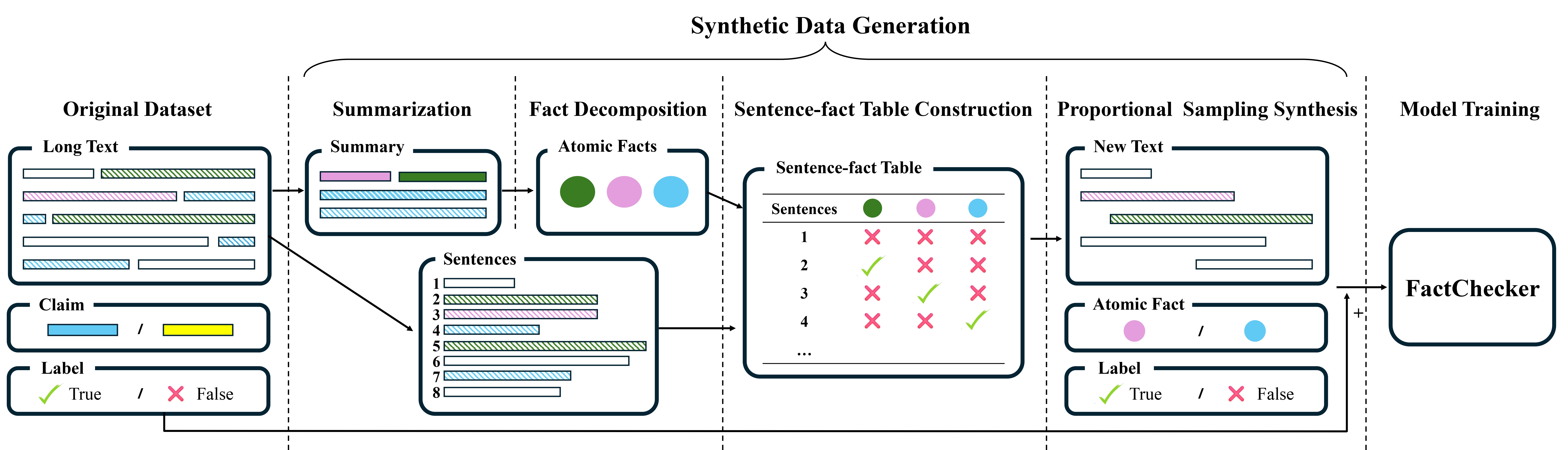}
\caption{The overview of \factchecker.}
%The process begins with document summarization, where the LLM generates the summary of the original text. This summary is then decomposed into atomic facts—basic, indivisible declarative sentences, each conveying a single piece of information. Following this, a sentence-fact table is constructed, where each row represents a sentence, and each column corresponds to an atomic fact, mapping the sentences to the facts they support. Using this table, we generate synthetic claim–text pairs by randomly selecting some of the sentences and an atomic fact, then assigning a veracity label based on whether the selected sentence supports the fact. Finally, these synthetic instances are combined with the original data to fine-tune a BERT-based fact-checking model, enhancing its ability of fact-checking.}
\label{fig:overview}
\end{figure}

In this study, we want to build a model, \factchecker, to decide whether each claim $c$ can be supported (true) or unsupported (false), according to a document $D$. To augment an existing dataset, we propose to use the original $D$ in the training data to construct a dataset of the text paired with claims with binary labels. In the synthetic data, the text is composed of sentences randomly selected from $D$; the claim corresponds to an atomic fact, and the label indicates whether the atomic fact is supported by the synthetic text.

Therefore, our proposed pipeline consists of 4 key steps (Figure~\ref{fig:overview}): (1) decomposing the original document to generate summaries and extract atomic facts, (2) constructing a sentence-fact table by mapping atomic facts to segmented sentences from the document, (3) generating synthetic data using the sentence-fact table, and (4) leveraging BERT-based models to train classifiers and make predictions.

\subsection{Generating synthetic data}

\paragraph{Document summarization}

%In our fact-checking tasks, documents are usually long texts including more than 20 sentences. A natural idea is to extract key information from long sentences by summarizing them. In our practice, we decompose the original document by extracting the document into a summary and then extracting atomic facts from the summary. The whole process will be completed by a large language model, which allows prompts as input to specify the instructions for each step. We compared different models and finally chose GPT-4 as the information extractor, considering its overall performance. We deployed the turbo-2024-04-09 version of the GPT-4 model (API version 2024-12-01-preview) through Azure OpenAI. This version allows us to give a prompt and receive a response from GPT-4. 

Our pipeline begins with a text summarization step. The primary goal of this stage is to generate a concise and coherent summary of each original document, which serves several purposes. First, summarization reduces the likelihood that a single fact is distributed across multiple sentences, thereby facilitating more precise and isolated extraction of atomic facts in subsequent steps. Second, by generating summaries distinct from the original documents, we introduce greater textual diversity, which can enhance the robustness and generalizability of downstream fact extraction and verification processes. This step ensures that the information is both condensed and reorganized, providing a clear foundation for the accurate identification of discrete facts. 

Here, we used the GPT-4 Turbo model (model version turbo-2024-04-09, API version 2024-12-01-preview) to generate the summary using the prompt in Figure~\ref{fig:prompt}A. 
We assume that these LLM-generated summaries are factually consistent with the original document \cite{tang2024minicheck}.

\begin{figure}[t]
\centering
\begin{subfigure}[b]{\textwidth}
\centering
\footnotesize
\begin{mdframed}[style=prompt]
\begin{lstlisting}
Document:[document]
Please generate a summary for the document with the following requirements: 
1. The summary should be a fluent and grammatical sentence.
2. The summary should be at least three sentences long.
3. The summary should cover information across the document.
Please respond **only** with a JSON object in the following format:
{
  "summary": "Your generated summary here."
}
\end{lstlisting}
\end{mdframed}
\vspace{-1em}
\caption{Prompt for summarization}
\end{subfigure}%

\begin{subfigure}[b]{\textwidth}
\centering
\begin{mdframed}[style=prompt]
\footnotesize
\begin{lstlisting}
Segment the sentence into individual facts as follows examples:
Example1,
Example2,
......
Sentence:[summary]
Please respond **only** with a JSON object in the following format:
{
  "facts": [
    "Fact 1",
    "Fact 2",
    "Fact 3",
    ...
  ]
}
\end{lstlisting}
\end{mdframed}
\vspace{-1em}
\caption{Prompt for fact extraction}
\end{subfigure}
\caption{Prompts in generating synthetic data}
\label{fig:prompt}
\end{figure}

\paragraph{Fact decomposition}

We then use the GPT-4 to decompose each sentence in the summary into a series of atomic facts. Following the definition in \cite{min2023factscore, tang2024minicheck}, an atomic fact is characterized as the most basic, indivisible declarative sentence, each conveying one piece of information.

%\yifan{I cannot find the prompt that asked for the most granular pieces of information.} 
We design prompts that use concrete fact-decomposition examples to instruct the model to identify and extract detailed information and present it in a structured, orderly format (Figure~\ref{fig:prompt}). This approach ensured that the resulting atomic facts were both precise and comprehensive, facilitating more reliable downstream processing.

\paragraph{Sentence-fact table construction}

%Based on the sentences and atomic facts of the original document, we can construct a sentence fact table. The rows are the sentences obtained after the long text in the document is segmented. The columns are atomic facts. Depending on the specific length of the original document, the shape of the table we generate may be slightly different. But considering that most common documents are long texts, the number of rows in the table we get is usually several times the number of columns.

%After obtaining this table, we use the entailment check method to assign values to the empty spaces in the table, following the following rules: The value of the i-th column and the j-th row depends on whether the j-th sentence can support the i-th fact. If it supports, it is assigned to 1, otherwise it is 0. The model used by the entailment check supports multiple choices. When we generate synthetic data, we rely on a large language model to make the judgment and assign value. In this step, we have a prior assumption that large language models can achieve good performance in the support verification task of short text and short text. This assumption is based on the fact that the capacity of existing large language models is not enough to achieve satisfactory performance in the task of long text and short text, or in other words, the cost of achieving good performance is relatively expensive. So we chose to use a large language model to handle simple tasks, that is, synthetic data, and then use a BERT-based model to complete the fact-checking task. 
After extracting individual atomic facts from the original documents, we organized the information into a structured sentence-fact table for each document. In this table, each row represents a sentence from the document, while each column corresponds to an atomic fact identified from the extraction process. 

We then performed an entailment check to verify the relationship between each sentence and each atomic fact, and populated the table with indicators (true or false) of whether each sentence entails a particular atomic fact. Notably, multiple sentences could entail the same fact. One sentence could entail multiple facts, too.

\paragraph{Proportional sampling synthesis}
\label{sec:syn}
%After getting the sentence fact table, we can generate synthetic data by sampling and recombining. The specific method is as follows: first select a certain number of sentences and splice them into a new text, and then select a fact as a new claim. As for the label of the new data, we judge it by the sum of the values of the row corresponding to the selected sentence under the corresponding column in the original table. If the sum is greater than 1, it proves that at least one sentence in the newly composed data can support the new claim. Therefore, we will consider the label of the newly composed data to be 1. Otherwise, it is considered to be 0.

%In this step, we assume that there are no sentences that can support a fact as a sum, but none of them can be considered to support this fact individually. As we said before, all facts are indivisible, so it is difficult to infer this fact by combining the information of several sentences instead of their individual information. This assumption ensures the rationality of the labels we give when synthesizing data.

%In this way, we can synthesize new data based on the sentence fact table. There are also some personalized generation options. For example, we can specify to select a certain proportion of sentences in the original document. It is also possible to generate multiple data by selecting multiple facts.

Next, we generated synthetic data using the sentence-fact tables. Specifically, $p\%$ sentences were randomly selected from each original document (referred to as ``Synthetic proportion'') and concatenated to form a new piece of text. A corresponding atomic fact was then chosen from the sentence-fact table to serve as a claim. The label for each synthetic instance was determined by referencing the sentence-fact table: if at least one of the selected sentences supported the chosen fact, the instance was labeled as ``true''; otherwise, it was labeled as ``false''.

\subsection{\factchecker development}

We adopted a BERT-based “pretrained encoder + classification head” architecture as our encoder~\cite{devlin2019bert}. The two input texts (a claim and its corresponding document) were concatenated into a single sequence formatted as ``\texttt{[CLS]} claim \texttt{[SEP]} document \texttt{[SEP]}''. This sequence was then fed through a BERT model, and the final‐layer hidden state of the \texttt{[CLS]}was used as the semantic representation of the pair. We then applied dropout before passing it through a single linear feedforward layer that output two logits. A softmax function was used to compute the probability. All parameters—including both the BERT encoder and the classification head were fine-tuned jointly by minimizing the cross-entropy loss on our binary labels.

\section{Experimental Settings}

\paragraph{Datasets}

\begin{table}[tbh]
\centering
\caption{Statistics of the data.}
\label{tab:data}
\begin{tabular}{lrrrrrrrrrr}
\toprule
Data & \multicolumn{3}{c}{Training} & \multicolumn{3}{c}{Test} & \multicolumn{2}{c}{Document}& \multicolumn{2}{c}{Claim}\\

\cmidrule(rl){2-4}\cmidrule(rl){5-7}\cmidrule(rl){8-9}\cmidrule(rl){10-11}
& Total & True & False & Total & True & False& Avg sent.& Avg word& Avg sent.& Avg word\\
\midrule
PubHealth & 5,863 & 4,101 & 1,762 & 987 & 599 & 388 & 20.42 & 511.26 & 1.05 & 12.54 \\
SciFact   & 957   & 616  & 341  & 338 & 216 & 122 & 10.15 & 245.83 & 1.00 & 12.19\\

% PubHealth & 5,863 & 4101 & 1762 & 987 & 599 & 388 & 20.42 & 12.54 \\
% SciFact   & 957   & 616  & 341  & 338 & 216 & 122 & 10.15 & 12.19\\
\bottomrule
\end{tabular}
\end{table}

To evaluate the performance of our overall model, we conducted experiments on two benchmark datasets: PubHealth and SciFact (Table~\ref{tab:data}).

%\textbf{Consider some data description. i.e. Average sentences in a document in both datasets we chose. Word count, etc.} 

\textbf{PubHealth} contains 9,832 document-claim pairs and their corresponding manually verified labels~\cite{kotonya-toni-2020-explainable-automated}. The labels contain four values: true, false, mixture, and unproven. 
Considering that our task mainly focused on binary classification, only true and false labeled instances were included.
To facilitate effective document decomposition, we also excluded documents that were either too short ($\leq 3$ sentences) or too long ($\geq 40$ sentences). Finally, we obtained a processed set of 5,863 instances for training and 987 instances for validation.
%In addition, the dataset also contains some other information, including the source of the document, the explanation, and the contributors of the manually verified labels.
%
%To better examine the document distribution, we use the "nltk.sent tokenize" function to split each document into sentences and observe the resulting distribution results. The average document contains 28.5 sentences (standard deviation = 20.7), and the number of sentences ranges from 1 to 432. The interquartile range ranges from 14 to 38 sentences, which indicates that a considerable number of documents are long and information-dense.
%
%In our implementation, we first performed a series of preprocessing on the data. We only retained the required columns, namely document, claim, and label. Then we dropped any empty values. In this task, the best way to deal with null values is to remove them, because filling null values is almost impossible. At the same time, 
%We removed documents that were too short ($\leq 3$ sentences) or too long ($\geq 40$ sentences) to ensure that our document decomposition step could proceed smoothly. After these processes, we finally got 5,863 instances for model development and 987 for validation.
%
In our subsequent experiments, we incrementally, randomly selected 500, 1,000, and 1,500 instances from the training dataset to generate synthetic data. To ensure the label balance, each subset comprised an equal number of instances from both labels. %The selection of data is random, but for comparison consideration, we ensured that the previous smaller dataset is a subset of the latter larger dataset, that is, the 1000 dataset contains the 500 dataset, and the same for 1500 and 1000.

%After constructing the training data, we did a similar pre-processing on the test data and got a finalized test data of 987 records.

\textbf{SciFact} is a collection of 1,409 claims verified against 5,183 abstracts~\cite{li2020paragraph-level}. For each claim, the corresponding abstracts were annotated as SUPPORT, CONTRADICT, or NEUTRAL with rationale sentences (also referred to as ``evidence'') from the abstracts. Here, we only picked the pairs with the label of SUPPORT  (true) and CONTRADICT (false).
Since the ground truth for the test set was not published, we used the standard training set (957 pairs) for training and the development set (338 pairs) for testing.

%consists of two main parts: a claims df table containing individual claims and their associated labels, and a corpus df table containing the corresponding documents (including titles and abstracts). They are linked by ids contained in both tables. The label aspect contains three values: SUPPORT, CONTRADICT, and null. Since our task focuses on binary classification problems, we remove all data with null labels. The processed training data contains 957 records. Considering that the test set from SciFact does not contain clear labels, we decided to use the validation set for testing which contains 338 records after similar pre-processing processes. 

\paragraph{Evaluation Metrics}

The precision, recall, and F1 scores were calculated.

%In order to objectively measure the performance of the binary classification model constructed in this study, this paper selects precision, recall and F1 score as core evaluation indicators. Precision represents the proportion of true positive examples in the samples determined by the model as positive, focusing on evaluating the accuracy of the prediction results; recall reflects the model's ability to detect all true positive examples, focusing on measuring the risk of omission; F1 score, as the harmonic average of the two, achieves a balance between accuracy and coverage, and can provide a more stable and robust overall performance evaluation in the context of uneven category distribution.

\paragraph{Model settings}
% The precision, recall, and F1 scores were calculated as evaluation m.etrics

All experiments were conducted on a Red Hat Enterprise Linux 8.10 (64-bit) system with dual Intel Xeon Gold 6226R CPUs (64 logical cores, 2.90 GHz) and two NVIDIA RTX A6000 GPUs, each with 48 GB of VRAM. 
%The CUDA version is 12.6, and the NVIDIA driver version is 560.35.03.
%
In our experiments, we fine-tuned the AllenAI SciBERT model (allenai/scibert\_scivocab\_uncased) using Hugging Face’s Transformers library. 
%We loaded the tokenizer and encoder via AutoTokenizer.from\_pretrained and AutoModel.from\_pretrained, producing a 768-dimensional pooled output corresponding to the [CLS] token. 
%We set freeze\_bert=False for better performance. 
All experiments were conducted using Python 3.11.11, PyTorch 2.5.1+cu121 (CUDA 12.1), Hugging Face Transformers 4.50.3, and Tokenizers 0.21.1. We fine-tuned the model for 10 epochs with a learning rate of $2\times10^{-5}$. The batch size was 16, and the AdamW optimizer with a weight decay of $1\times10^{-2}$ is employed. 

%We used nn.BCEWithLogitsLoss() on the raw logits against float labels. A linear scheduler without warmup (get\_linear\_schedule\_with\_warmup with num\_warmup\_steps=0) decayed the learning rate to zero over the total training steps, computed as (len(train\_loader) // accumulation\_steps) × epochs. Inputs were truncated or padded to a maximum length of 512 tokens, with a per-device batch size of 16 and gradient accumulation over 2 steps (yielding an effective batch size of 32). Data loading used five worker processes, and, when multiple GPUs were detected, the model was wrapped in nn.DataParallel for multi-GPU training.

\section{Results and Discussion}

\subsection{Results on PubHealth and SciFact}

Table~\ref{tab:pubhealth} shows the results on the PubHealth dataset, where column ``synthetic proportion'' suggests the proportion of sentences selected from the original documents (Section~\ref{sec:syn}).
%, we constructed three subsets to test our pipeline performance. They are 500, 1000, 1500 subsets containing corresponding number of original data. 
%Also, we have a parameter to determine to what proportion we pick in the sentences from the original documents. 
We combined the synthetic data with the original training set to form an augmented training set. Note that when the proportion is 0\%, no synthetic data was incorporated. 
In this case, the model was trained solely on the original dataset, which serves as the baseline for comparison.
We then compared the model's performance when trained with the augmented dataset versus using only the original training data.
Table~\ref{tab:pubhealth} shows that the best proportion in each subset all achieves higher performance compared with only using the original training data. 

On the subset with 500 instances, the best-performing system was trained using 90\% of sentences from the original documents to construct the synthetic data, achieving an F1 score of 0.792. Out of 10 experimental settings, four outperformed the baseline. These results indicate that the effectiveness of synthesized data in enhancing the \factchecker’s performance depends on selecting an appropriate proportion that provides meaningful and informative training examples.

%Then, randomly pick one fact from each sentence-fact table as the claim. The label is given by summing up the chosen sentences and facts' corresponding values in the table, and is judged based on whether the sum is 0 or not. This system gives a 0.792 F1 score, which is significantly higher than only using the original training data with an F1 score of 0.776.

A similar trend is observed across larger subsets. On the subset with 1,000 instances, employing just 20\% of the original sentences led to an impressive F1 score of 0.806, notably higher than the baseline of 0.792.  Here, 3 out of the 10 experimental settings overperformed the baseline. The results are even more compelling on the 1,500-instance subset: selecting only 10\% of sentences yielded the highest F1 score of 0.831, compared to the baseline of 0.812. In this case, a remarkable 6 out of 10 settings surpassed the baseline, underscoring the robust potential of fine-tuned proportions for maximizing system performance.
%, achieving an F1 score of 0.806. One 1500 subsets, 10\% is the best choice, achieving 0.8305 F1 score, outperforming 0.8119 by original data only.

Table~\ref{tab:pubhealth} also shows the results on the SciFact dataset. For most proportions of synthetic data, models' performance surpassed that of those fine-tuned with only the original data. Notably, the highest F1 score was observed when the proportion was set to 100\%, reaching 0.792 -- substantially higher than the 0.741 achieved using only the original dataset.

\begin{table}[tbh]
\centering
\caption{Results on PubHealth and SciFact.}
\label{tab:pubhealth}
\begin{tabular}{@{}lrrrrrrrrrrrr@{}}
\toprule
\multirow{3}{*}{Synthetic proportion} & \multicolumn{9}{c}{PubHealth} & \multicolumn{3}{c}{SciFact}\\
\cmidrule(rl){2-10}\cmidrule(rl){11-13}
 & \multicolumn{3}{c}{500} & \multicolumn{3}{c}{1,000} & \multicolumn{3}{c}{1,500}\\
\cmidrule(rl){2-4}\cmidrule(rl){5-7}\cmidrule(rl){8-10}
 & P & R & F  & P & R & F & P & R & F & P & R & F\\
\midrule
0\% (baseline) & 0.949 & 0.656 & 0.776  & 0.960 & 0.674 & 0.792 & 0.859 & 0.728 & 0.812 & 0.716 & 0.769 & 0.741 \\
10\% & 0.964 & 0.626 & 0.759 & 0.963 & 0.658 & 0.782  & 0.891 & 0.780 & \textcolor{red}{0.831} & 0.718 & 0.712 & 0.714 \\
20\% & 0.903 & 0.621 & 0.736  & 0.849 & 0.768 & \textcolor{red}{0.806}  & 0.912 & 0.741 & 0.813 & 0.776 & 0.686 & 0.727 \\
30\% & 0.842 & 0.723 & 0.778  & 0.904 & 0.708 & 0.794  & 0.902 & 0.746 & 0.815 & 0.741 & 0.738 & 0.737 \\
40\% & 0.842 & 0.728 & 0.781  & 0.869 & 0.711 & 0.782  & 0.911 & 0.732 & 0.811 & 0.745 & 0.759 & 0.752 \\
50\% & 0.741 & 0.780 & 0.760  & 0.838 & 0.733 & 0.782  & 0.892 & 0.744 & 0.811 & 0.687 & 0.832 & 0.751 \\
60\% & 0.812 & 0.765 & 0.788  & 0.938 & 0.628 & 0.752  & 0.888 & 0.743 & 0.808 & 0.672 & 0.925 & 0.775 \\
70\% & 0.686 & 0.838 & 0.754  & 0.950 & 0.664 & 0.782  & 0.882 & 0.771 & 0.822 & 0.658 & 0.985 & 0.789 \\
80\% & 0.728 & 0.798 & 0.761  & 0.909 & 0.664 & 0.767  & 0.855 & 0.770 & 0.810 & 0.641 & 0.984 & 0.777 \\
90\% & 0.843 & 0.746 & \textcolor{red}{0.792}  & 0.896 & 0.720 & 0.798  & 0.833 & 0.796 & 0.813 & 0.648 & 0.986 & 0.782 \\
100\% & 0.664 & 0.917 & 0.770  & 0.948 & 0.666 & 0.782  & 0.810 & 0.825 & 0.816 & 0.666 & 0.979 & \textcolor{red}{0.792} \\

\bottomrule
\end{tabular}
\end{table}

%\subsection{Results for SciFact}

% \begin{table}[]
% \centering
% \begin{tabular}{lrrr}
% \toprule
% Proportion & P & R & F \\
% \midrule
% 10\% & 0.718 & 0.712 & 0.714 \\
% 20\% & 0.776 & 0.686 & 0.727 \\
% 30\% & 0.741 & 0.738 & 0.737 \\
% 40\% & 0.745 & 0.759 & 0.752 \\
% 50\% & 0.687 & 0.832 & 0.751 \\
% 60\% & 0.672 & 0.925 & 0.775 \\
% 70\% & 0.658 & 0.985 & 0.789 \\
% 80\% & 0.641 & 0.984 & 0.777 \\
% 90\% & 0.648 & 0.986 & 0.782 \\
% 100\% & 0.666 & 0.979 & 0.792 \\
% official & 0.716 & 0.769 & 0.741 \\
% \bottomrule
% \end{tabular}
% \caption{Results on SciFact}
% \label{tab:SciFact}
% \end{table}This suggests that 
% \yifan{Add one more sentence to discuss what this observation suggests.}

\subsection{A pilot study in detecting hallucinations}

To demonstrate the effectiveness of \factchecker, we conducted a pilot study focused on detecting hallucinations in LLM-generated text summaries. For each summary, we constructed a sentence-fact table using both the original document and the corresponding LLM-generated summary, following a similar approach as described in Section~\ref{sec:syn}. Our \factchecker was then applied to populate the table with values (true or false), indicating whether each sentence supported a given decomposed fact. If all the values in a column were 0, we marked that column as abnormal, suggesting that the corresponding fact may not be supported by any sentence in the original document. Such cases likely indicate hallucinations produced by LLM. Finally, we verified these cases manually.

In this pilot study, we examined 500 instances from the PubHealth dataset that were not included in our earlier experiments. We used our BERT-based \factchecker trained on the 1,500 subsets to assign values in the sentence-fact tables for these new cases. This process revealed two abnormal instances. In the first case (Figure~\ref{fig:first case}), the abnormality was caused by hallucination, in which LLM appeared to use external knowledge or fabricate facts not present in the source. 
In the second case (Figure~\ref{fig:second case}), the abnormal fact identified in the summary could only be inferred by combining information from several sentences, but was not directly supported by any single sentence. 
The first scenario is an ideal demonstration for our pilot study, whereas the second one highlights a potential future research direction. 

\begin{figure}[t]
\begin{mdframed}[style=prompt]
\footnotesize 

\textbf{Document:}   A somewhat disturbing image of an unknown animal’s feet has been circulating online for several years, alongside a variety of claims about its origins. In March 2018, the photograph was re-purposed and shared as if the creature whose prints it captured had been spotted recently on a game trail in Cleburne County, Alabama:  We have not been able to identify the source of this photograph. However, the claims made in recent Facebook posts about it are false: This picture was not taken in March 2018, and the animal it captured traces of (if any)  did not destroy a trail camera or kill an Alabama farmer’s livestock. This image has been online since at least as far back as October 2015, when it was shared on Instagram along with a story about a sailor in Oman who was “shocked” to find the image on his camera. This photograph has also been used to illustrate similar stories about a strange creature that was supposedly found in India, Brazil, and Hawaii. Needless to say, we’re skeptical that this exact same photograph was pulled from cameras in so many different locations and times. So what does this picture actually show? While this image is frequently claimed to depict evidence of a lobisomem, South America’s version of a werewolf, we’re fairly certain that this is a dog (or another extant species) suffering from mange: “Mange (demodicosis) is an inflammatory disease in dogs caused by various types of the Demodex mite. When the number of mites inhabiting the hair follicles and skin of the dog become exorbitant, it can lead to skin lesions, genetic disorders, problems with the immune system and hair loss (alopecia).” In fact, the paws in this viral photograph closely resemble a 2014 photograph taken by the School of Veterinary Medicine and Science at the University of Nottingham, UK, of a dog suffering from a severe case of sarcoptic mange:

\vspace{1em}

\textbf{Summary:} A mysterious image of an unknown animal's feet has been circulating online, with various claims about its origins, but experts have debunked these claims, concluding that the image likely shows a dog suffering from mange, \textcolor{red}{a skin disease caused by Demodex mites.}

\end{mdframed}
%\vspace{-1em}
\caption{One example illustrates that the abnormality (highlighted in red) resulted from a hallucination.}
\label{fig:first case}
\end{figure}

\begin{figure}[t]
\begin{mdframed}[style=prompt]
\footnotesize
\textbf{Document:}   In February 2015, the web site Cronica MX published an article reporting that former Chicago Bull basketball star Michael Jordan had passed away of a heart attacj at the age of 52:  This morning the ex basket ball player and also icon of world sports, Michael Jordan, was found dead while in his sleep in his residence in North Caroline. Around 3 am eastern time, Yvette Prieto, now widow of the star noticed that her husband showed no vital signs and she proceeded to call 911. Minutes after the paramedics arrived, they confirmed the death of the athlete due to a heart attack. The wife, Yvette Prieto, said she feels devastated for the sudden death of his beloved husband and in a short speech to some news reporters she thanked them for all the support. She said that Jordan was a great person and everybody loved him. The web site also produced a video in order to lend some credibility to the claim that His Airness had died. Although the clip was designed to look like a breaking news segment, the footage posted to YouTube was not anything broadcast by a credible news network:  \textcolor{red}{The dead giveaway that this video was a fake (and the report of Jordan’s death a hoax)} came at the 25-second mark, when reporter Rich Eisen appeared on screen to give a tearful goodbye to the sports legend. Although the footage of Eisen was real, the sports reporter was not shown mourning the death of Michael Jordan; rather, \textcolor{red}{that portion of the clip was a recycled portion of an NFL Game Day broadcast} from January 2015 which aired shortly after the death of Eisen’s fellow ESPN anchor Stuart Scott:

\vspace{1em}

\textbf{Summary:} In February 2015, a false report spread that Michael Jordan had passed away at the age of 52 due to a heart attack, but the claim was debunked when a video posted on YouTube was revealed to be a fake, \textcolor{red}{featuring a recycled segment from an NFL Game Day broadcast} and a tearful goodbye from reporter Rich Eisen, which was actually a tribute to his colleague Stuart Scott, not Michael Jordan. The first situation is a perfect use case for our pilot study, while the second one indicates a potential future research topic. 
\end{mdframed}
%\vspace{-1em}
\caption{One example illustrates that the abnormality (highlighted in red) could only be inferred by combining information from several sentences.}
\label{fig:second case}
\end{figure}

\subsection{Discussion}

%\paragraph{Impact of data size and sentence selection on \factchecker Performance}

Based on the results obtained from Table~\ref{tab:pubhealth}, several general patterns can be identified. 
First, we found that increasing the amount of data allows the \factchecker to capture more knowledge and, consequently, achieve better performance. The best results were achieved with F1 scores of 0.792, 0.806, and 0.831 for the 500, 1,000, and 1,500 subsets, respectively. Second, the optimal proportion of sentences selected from the original documents varies across subsets, indicating that there is no universally optimal selection strategy. Thus, further investigations are needed to establish a general model that consistently uses a fixed proportion for all cases.

%In addition to our main findings, we also had some other discoveries and insights when constructing the sentence-fact table.

%\paragraph{Scalable synthetic data generation from text}

While we only experimented with several settings in this study, it is worth noting that our proposed data synthesis procedure can be repeated multiple times to efficiently generate a large volume of synthetic data. Furthermore, adjusting the proportion of sentences retained from the original documents could further enhance the flexibility and diversity of this synthetic data generation pipeline.

This study has several limitations. First, our study mainly focused on the two datasets, PubHealth and SciFact, which constrain the generalizability of this study. Second, in our experiment, a single BERT-based \factchecker is employed. In our future work, more BERT models can be tried to discover more innovative results.

\section{Conclusion}
In this study, we proposed an LLM-driven synthetic data generation pipeline. The sentence-fact table generated through this pipeline, along with the resulting synthetic data, brings new insights into fact-checking tasks. Firstly, the synthetic data intuitively helps solve the insufficiency in training data. Meanwhile, the experiments on the two benchmark datasets with our \factchecker also provide meaningful observations. Based on this pipeline, we further show that it is possible to fine-tune a smaller model to effectively verify facts within grounding documents and detect hallucinations in LLM-generated text. In conclusion, the novelty proposed in this study provides a potential feasible solution and future development direction for the field of fact-checking.
%ournovel  is proposedbased on andsubsequent the related topics of we can achieve better performance for the same fact-checker model.

\vspace{1em}
\noindent
\textbf{Acknowledgment}: This project was sponsored by the National Library of Medicine grants R01LM014344, R01LM014306, and R01LM014573.

\setlength{\bibsep}{3pt plus 0.3ex}
\bibliographystyle{unsrtnat}
\bibliography{ref}

\begin{thebibliography}{10}
\providecommand{\natexlab}[1]{#1}
\providecommand{\url}[1]{\texttt{#1}}
\expandafter\ifx\csname urlstyle\endcsname\relax
  \providecommand{\doi}[1]{doi: #1}\else
  \providecommand{\doi}{doi: \begingroup \urlstyle{rm}\Url}\fi

\bibitem[Kotonya and Toni(2020)]{kotonya-toni-2020-explainable-automated}
Neema Kotonya and Francesca Toni.
\newblock Explainable automated fact-checking for public health claims.
\newblock In \emph{Proceedings of the 2020 Conference on Empirical Methods in Natural Language Processing (EMNLP)}, pages 7740--7754, Online, 2020. Association for Computational Linguistics.
\newblock \doi{10.18653/v1/2020.emnlp-main.623}.

\bibitem[Vlachos and Riedel(2014)]{vlachos-riedel-2014-fact}
Andreas Vlachos and Sebastian Riedel.
\newblock Fact checking: Task definition and dataset construction.
\newblock In Cristian Danescu-Niculescu-Mizil, Jacob Eisenstein, Kathleen McKeown, and Noah~A. Smith, editors, \emph{Proceedings of the ACL 2014 Workshop on Language Technologies and Computational Social Science}, pages 18--22, Baltimore, MD, USA, June 2014. Association for Computational Linguistics.
\newblock \doi{10.3115/v1/W14-2508}.
\newblock URL \url{https://aclanthology.org/W14-2508/}.

\bibitem[Waszak et~al.(2018)Waszak, Kasprzycka-Waszak, and Kubanek]{waszak2018spread}
Przemyslaw~M. Waszak, Wioleta Kasprzycka-Waszak, and Alicja Kubanek.
\newblock {The Spread of Medical Fake News in Social Media – The Pilot Quantitative Study}.
\newblock \emph{Health Policy and Technology}, 7\penalty0 (2):\penalty0 115--118, 2018.
\newblock \doi{10.1016/j.hlpt.2018.03.002}.

\bibitem[Kazari et~al.(2025)Kazari, Chen, and Shakeri]{kazari2025scaling}
Kamyar Kazari, Yong Chen, and Zahra Shakeri.
\newblock Scaling public health text annotation: Zero‐shot learning vs. crowdsourcing for improved efficiency and labeling accuracy, 2025.
\newblock URL \url{https://arxiv.org/abs/2502.06150}.

\bibitem[Thorne et~al.(2018)Thorne, Vlachos, Christodoulopoulos, and Mittal]{thorne2018fever}
James Thorne, Andreas Vlachos, Christos Christodoulopoulos, and Arpit Mittal.
\newblock {FEVER: A Large-Scale Dataset for Fact Extraction and VERification}.
\newblock In \emph{Proceedings of the 2018 Conference of the North American Chapter of the Association for Computational Linguistics: Human Language Technologies}, pages 809--819, 2018.
\newblock \doi{10.18653/v1/N18-1074}.

\bibitem[Wadden et~al.(2020)Wadden, Lin, Lo, Wang, van Zuylen, Cohan, and Hajishirzi]{wadden-etal-2020-fact}
David Wadden, Shanchuan Lin, Kyle Lo, Lucy~Lu Wang, Madeleine van Zuylen, Arman Cohan, and Hannaneh Hajishirzi.
\newblock Fact or fiction: Verifying scientific claims.
\newblock In Bonnie Webber, Trevor Cohn, Yulan He, and Yang Liu, editors, \emph{Proceedings of the 2020 Conference on Empirical Methods in Natural Language Processing (EMNLP)}, pages 7534--7550, Online, November 2020. Association for Computational Linguistics.
\newblock \doi{10.18653/v1/2020.emnlp-main.609}.
\newblock URL \url{https://aclanthology.org/2020.emnlp-main.609/}.

\bibitem[Tang et~al.(2024)Tang, Laban, and Durrett]{tang2024minicheck}
Liyan Tang, Philippe Laban, and Greg Durrett.
\newblock {MiniCheck}: Efficient fact-checking of {LLMs} on grounding documents.
\newblock In \emph{Proceedings of the 2024 Conference on Empirical Methods in Natural Language Processing}, pages 8818--8847, Stroudsburg, PA, USA, 16~April 2024. Association for Computational Linguistics.
\newblock \doi{10.18653/v1/2024.emnlp-main.499}.

\bibitem[Min et~al.(2023)Min, Krishna, Lyu, Lewis, Yih, Koh, Iyyer, Zettlemoyer, and Hajishirzi]{min2023factscore}
Sewon Min, Kalpesh Krishna, Xinxi Lyu, Mike Lewis, Wen-Tau Yih, Pang Koh, Mohit Iyyer, Luke Zettlemoyer, and Hannaneh Hajishirzi.
\newblock {FActScore}: Fine-grained atomic evaluation of factual precision in long form text generation.
\newblock In \emph{Proceedings of the 2023 Conference on Empirical Methods in Natural Language Processing}, page 12076–12100, Stroudsburg, PA, USA, 2023. Association for Computational Linguistics.
\newblock \doi{10.18653/v1/2023.emnlp-main.741}.

\bibitem[Devlin et~al.(2019)Devlin, Chang, Lee, and Toutanova]{devlin2019bert}
Jacob Devlin, Ming-Wei Chang, Kenton Lee, and Kristina Toutanova.
\newblock Bert: Pre-training of deep bidirectional transformers for language understanding.
\newblock In \emph{Proceedings of the 2019 Conference of the North American Chapter of the Association for Computational Linguistics: Human Language Technologies}, pages 4171--4186. ACL, 2019.

\bibitem[Li et~al.(2020)Li, Burns, and Peng]{li2020paragraph-level}
Xiangci Li, Gully Burns, and Nanyun Peng.
\newblock A paragraph-level multi-task learning model for scientific fact-verification.
\newblock \emph{arXiv [cs.CL]}, 28~December 2020.

\end{thebibliography}

\end{document}